\documentclass[a4paper, 10pt, conference]{ieeeconf}      

\IEEEoverridecommandlockouts                              
\usepackage[utf8]{inputenc}
\usepackage[T1]{fontenc}
\usepackage[pdftex]{graphicx}
\usepackage{graphicx}
\usepackage{tikz}
\usepackage{subfigure} 
\graphicspath{{../Figures/}}
\usepackage{booktabs} 
\usepackage{multirow} 
\usepackage{threeparttable} 
\usepackage{float}
\usepackage{multicol}
\usepackage{color, colortbl}
\usepackage[]{algorithm2e}
\usepackage{amsmath} 
\usepackage{array}  
   
\usepackage{bm} 

\title{\LARGE \bf
Probabilistic Egocentric Motion Correction of Lidar Point Cloud and Projection to Camera Images for Moving Platforms
}

\author{Mao Shan, Julie Stephany Berrio,  Stewart Worrall, Eduardo Nebot 
\thanks{The authors  are with the Australian Centre for Field Robotics at the University of Sydney, NSW, Australia.
       E-mails: {\tt\small \{m.shan, j.berrio, s.worrall, e.nebot\}@acfr.usyd.edu.au}}
}

\begin{document}

\maketitle
\thispagestyle{empty}
\pagestyle{empty}

\begin{abstract}
The fusion of sensor data from heterogeneous sensors is crucial for robust perception in various robotics applications that involve moving platforms, for instance, autonomous vehicle navigation. In particular, combining camera and lidar sensors enables the projection of precise range information of the surrounding environment onto visual images. It also makes it possible to label each lidar point with visual segmentation/classification results for 3D mapping, which facilitates a higher level understanding of the scene. The task is however considered non-trivial due to intrinsic and extrinsic sensor calibration, and the distortion of lidar points resulting from the ego-motion of the platform. Despite the existence of many lidar ego-motion correction methods, the errors in the correction process due to uncertainty in ego-motion estimation are not possible to remove completely. It is thus essential to consider the problem a probabilistic process where the ego-motion estimation uncertainty is modelled and considered consistently. The paper investigates the probabilistic lidar ego-motion correction and lidar-to-camera projection, where both the uncertainty in the ego-motion estimation and time jitter in sensory measurements are incorporated. The proposed approach is validated both in simulation and using real-world data collected from an electric vehicle retrofitted with wide-angle cameras and a 16-beam scanning lidar.
\end{abstract}

\section{Introduction}

\begin{figure}[!t]
	\centering
	\subfigure[]{ 
		\label{fig:LidarProjectionDetails:a}
		\includegraphics[height=2.6in]{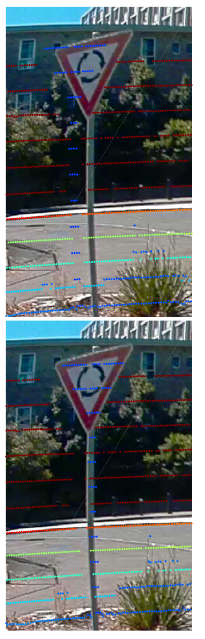}}
	\hspace{-0.19in}
	\subfigure[]{ 
		\label{fig:LidarProjectionDetails:b}
		\includegraphics[height=2.6in]{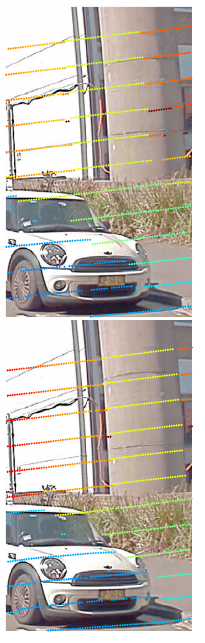}}
	\hspace{-0.19in}
	\subfigure[]{ 
		\label{fig:LidarProjectionDetails:c}
		\includegraphics[height=2.6in]{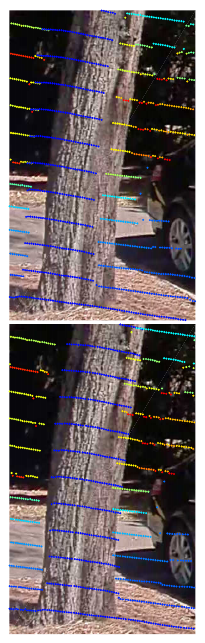}}
	\hspace{-0.19in}
	\subfigure[]{ 
		\label{fig:LidarProjectionDetails:d}
		\includegraphics[height=2.6in]{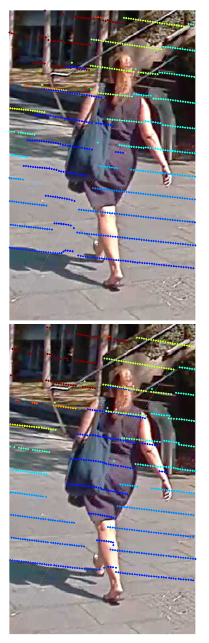}}
	\caption{Examples of projecting lidar points to images before (top row) and after (bottom row) ego-motion correction. Misalignment of lidar points and features in the images due to ego-motion distortion can be clearly seen for a traffic sign in (a), a parked car and a building behind in (b), a tree trunk in (c), and lastly a walking pedestrian in (d). The figures in the bottom row show the improved results in the projection of ego-motion corrected lidar points using the proposed approach. The ego-motion correction uncertainty is available for each projected point, but not shown in the figures due to size constraints. The projected lidar points are colour-coded by range.}
	\label{fig:LidarProjectionDetails} 
\end{figure}

To navigate through any environment, a mobile robot platform is required to perceive the environment and achieve some level of understanding of the surroundings. In many sophisticated systems, this requires the combining of information from heterogeneous on-board sensors.
Lidars and cameras are complementary sensors that are extensively used in various robotic systems. Each sensor has a different strength---lidars offer precise range and reflective intensity measurements that are registered in 3D space, while cameras provide rich information of colour, texture and other visual features only in 2D. A considerable proportion of autonomous driving solutions proposed and developed by the automotive industry and research institutes rely on multiple cameras and lidars, in particular multi-beam lidars, to capture the activity of road users in the vicinity and to build contextual information---pedestrians, cyclists, other vehicles, traffic signs, lane markings, the road itself, etc.---in a traffic scene. Through the fusion of information from the two sensors of different modalities, we are able to transfer relevant data from the lidar to the camera domain, and vice versa, providing a better understanding of the surroundings' structure \cite{paper:ChienKlette2016}. It is thus of great importance to achieve accurate and robust perception by fusing camera and lidar information in a consistent manner.

Although often over-simplified in many applications as being measured at a single time, each point contained in a lidar scan is in fact captured at a slightly different time due to the laser firing cycles. The motion of the egocentric robot platform causes distortion in the lidar measurements as the sensor coordinate system moves along with the platform during the period of a scan. In theory, every 3D point is measured from a temporally unique frame of reference. The lidar points therefore must be compensated for ego-motion and transformed into a common reference coordinate system before further point cloud and sensor fusion related processing can take place. This can include, for instance, lidar feature extraction, projection to image frames, transferring segmentation results from the image space onto the 3D point cloud, or 3D mapping. The ego-motion correction becomes more essential for higher speed motion of the robot system, where the distortion caused by ego-motion tends to be more severe. Examples presented in Figure \ref{fig:LidarProjectionDetails} illustrate misalignment of lidar and visual features in the environment when projecting uncorrected lidar points to images, which can cause degraded performance in the sensor fusion. Interested readers can refer to \cite{paper:RiekenMaurer2016} for quantitative analyses of the time-related effects of moving scanning sensors on different perception tasks for multiple sensor systems.



Depending on the way the underlying ego-motion estimation of the sensor platform is conducted, there are two main categories of existing approaches proposed in the literature to correct the 3D lidar point cloud distortion due to ego-motion of the platform. In the first type, such as \cite{paper:SchneiderHimmelsbach2010, paper:HimmelsbachMuller2010, paper:MerriauxDupuis2017}, lidar scans are corrected by exploiting information from motion estimation sensors, such as IMU and odometry measurements. More complicated work presented in \cite{paper:VargaCostea2017, paper:ByunNa2015} obtains vehicle pose translation and rotation by fusing precise GNSS and IMU measurements. 
However, high end GNSS/INS units are costly, 
and their desired performance would not be achieved in GNSS denied environments. The other type relies on lidar based odometry estimation \cite{paper:TangYoon2018}, which eliminates the requirement of additional hardware. It can be further decomposed to simultaneous localization and mapping (SLAM) based approaches \cite{paper:MoosmannStiller2011, paper:ZhangSingh2014} that estimate the ego-motion by comparing point cloud features, and iterative closest point (ICP) based methods \cite{paper:HongKo2010} which infer the ego-motion through matching of consecutive scans. The SLAM based approaches are preferred in 3D map construction, yet loop closure is not achievable in some environments. ICP based methods are prone to errors brought by moving objects in the scene, such as pedestrians and vehicles \cite{paper:VargaCostea2017}.
Overall, it requires a substantial computational overhead for extracting features from a lidar point cloud and comparing lidar scans in these approaches.


None of the above approaches provides a way to estimate the uncertainty associated with each of the 3D lidar points and/or 2D image points in ego-motion correction process. We stress that there is always some uncertainty in 3D space associated with each motion corrected lidar point brought by errors in ego-motion estimation, regardless of which odometry sensors and/or estimation frameworks are adopted. Likewise, the motion corrected points when projected into a camera coordinate system also contain uncertainty in 2D image coordinates. The uncertainty is often considerable under many circumstances and has to be estimated along with the ego-motion correction process. Thus, the perception system can benefit from the uncertainty estimates in the subsequent point cloud and sensor fusion processing pipeline.

A probabilistic approach was proposed by \cite{paper:LeGentilVidalCalleja2018} that includes the correction due to motion distortion in 3D point clouds using IMU data considering measurement uncertainty. Yet, the approach mainly focuses on recovery of extrinsic calibration parameters of a lidar-IMU tightly coupled system, which does not produce explicit estimation uncertainty for corrected lidar points. The more recent work \cite{paper:CharikaShan2019} presents a probabilistic approach to estimate the uncertainty in the lidar-to-camera projection process. It employs a Jacobian based uncertainty model to estimate for each projected lidar point the combined uncertainty (in 2D) resulting from noise in ego-motion correction and errors in sensors' extrinsic and intrinsic calibration. Nevertheless, the uncertainty estimation for ego-motion corrected lidar points themselves (in 3D) is not supported by the approach.

The paper examines probabilistic ego-motion correction of lidar 3D point clouds to an arbitrary reference timestamp and projection to 2D image frames considering uncertainty in ego-motion estimation of the moving platform. On top of ego-motion correction outcomes, the proposed approach provides uncertainty estimates separately for each ego-motion corrected 3D lidar point and each projected 2D pixel point. Besides, the proposed approach considers additional uncertainty brought by time jitter in sensor data timestamps, which is a practical issue in many robotics systems. The proposed approach is validated using real-world data collected on an electric vehicle platform. Simulation results are also presented to quantitatively evaluate the proposed approach and assess the estimator credibility.

The remainder of the paper is organised as follows. Section \ref{sec:approach} presents the details of the proposed approach, including the probabilistic lidar ego-motion correction and the projection to an image frame. The experiment outcomes are presented in Section \ref{sec:results}, where simulation results are also included. Lastly, Section \ref{sec:conlusions} concludes the paper.

\section{The Proposed Approach}
\label{sec:approach}

In this section, we elaborate on the calibration of multiple cameras and a lidar in our experimental platform, and the methodology to estimate uncertainties as a result of probabilistic lidar ego-motion correction and projection.





\subsection{Calibration}

The cameras located on the electric vehicle platform used in the experiment have a lens of $100^\circ$ horizontal field of view, which is classified as a fisheye lens. We have calibrated the cameras by using a variation of the ROS package \textit{camera\_calibration} \cite{ros_camera_calib} that uses a generic camera model \cite{fish_eye_model}. The camera intrinsic parameters for this model consist of the focal length, principal points and 4 fisheye equidistant distortion coefficients. These values are critical for lidar-to-camera projection and the subsequent sensor fusion.

The extrinsic camera calibration is challenging when working with wide angle cameras due to the significant distortions in the lens. The extrinsic calibration in our electric vehicle platform was conducted as specified in the previous work \cite{SurabhiITSC}. This process uses a checkerboard from which the same features are extracted by both the camera and the lidar. The features are the centre point and the normal vector of the board. These features are fed to a genetic algorithm which is in charge of optimising the geometrical extrinsic parameters of the 3D transformation \(T_{cam}^{ld}\) between the two sensors.

\subsection{Probabilistic Lidar Ego-Motion Correction and Projection to Image Frame}




Usually within a lidar scan, lidar measurements with similar timestamps are grouped into a single lidar packet, with a common timestamp assigned to the grouping for convenience of processing. For instance, the Velodyne VLP-16 software driver used in our electric vehicle platform produces 76 packets for each full revolution scan. Each packet covers an azimuth angle of approximately \(4.74^{\circ}\). Alternatively, processing can be based on each individual lidar point with its own precise timestamp, though this comes at a significantly higher computational cost.

\begin{figure}[!t]
\vspace{2mm}
\centerline{
\includegraphics[width=0.9\columnwidth]{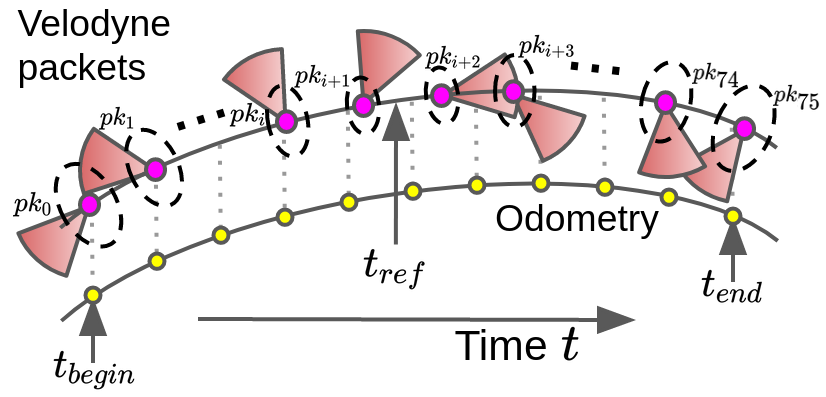}
}
\caption{Lidar point cloud motion correction process.}
\label{fig:time_line}
\end{figure}

Each of the lidar packets is transformed based on the estimated delta translation and rotation of the vehicle platform between the packet's timestamp and the reference timestamp \(t_{ref}\), as illustrated in Figure \ref{fig:time_line}. The proposed approach makes use of the unscented transform (UT) to propagate the uncertainties from the ego-motion estimation to corrected 3D lidar points and then to projected pixel coordinates in each camera image. The entire process can be divided into three cascaded stages, namely vehicle ego-motion estimation, lidar motion correction, and lidar-to-camera projection, each can be fitted into the UT pipeline.

The reference time \(t_{ref}\) is usually chosen to be the timestamp corresponding to a common frame of reference where sensor fusion or subsequent processing happens. In scenarios where camera-lidar sensor fusion is desired, rectification of the lidar points have to be matched with the timestamp of the associated camera frame before the lidar-to-camera projection can be carried out \cite{paper:VargaCostea2017}. For instance, the \(t_{ref}\) can be set to coincide with the timestamp of the most recent or closest image.

We assume a lidar scan is comprised of a set of \(N\) packets and their timestamps denoted as
\begin{equation}
\left\{pk_{i}, t_{i}^{lpk}\right\}_{i=0}^{N-1},
\end{equation}
where \(pk_{i}\) contains a set of \(M\) 3D lidar measurement points \(\left\{\bm{z}_{i,j}^{ld}\right\}_{j=0}^{M-1}\), and \(\bm{z}^{ld} = \begin{bmatrix} x^{ld} & y^{ld} & z^{ld} & 1 \end{bmatrix}^{T}\).

Before we proceed, the UT state decomposition and recovery functions are presented in Table \ref{table:UT_Decompose} and Table \ref{table:UT_Restore} respectively for the convenience of subsequent discussion, where \(\lambda = \alpha^2\left(d+\kappa\right)-d\), \(d = dim\left(\textbf{x}\right)\) is the dimension of state \(\textbf{x}\), scaling parameters \(\kappa \ge 0\), \(\alpha \in \left(0, 1\right]\), and \(\beta = 2\) for Gaussian distribution, \(\left(\sqrt{\bm{\Sigma}}\right)_i\) is to obtain the \(i^{th}\) column of the matrix square root \(\textbf{R} = \sqrt{\bm{\Sigma}}\), which can be computed by Cholesky decomposition such that we have \(\bm{\Sigma} = \textbf{R}\textbf{R}^{T}\).

\subsubsection{Vehicle Ego-Motion Estimation}

A proper way to estimate the ego-motion of the moving sensor platform is required to address its changing poses when perceiving the environment. In the electric vehicle platform used in our experiments, instantaneous linear and angular velocities are read from onboard wheel encoders and an IMU, respectively, at a rate of 100 Hz. Based on a sequence of monotonically increasing packet timestamps \(\bm{t}_{0:N-1}^{pk} = \left\{t_{i}^{pk}\right\}_{i=0}^{N-1}\), it is reasonable to construct a sequence of linear velocity vectors \(\bm{z}_{0:N-1}^{v} = \left\{\bm{z}_{i}^{v}\right\}_{i=0}^{N-1}\) corresponding to \(\bm{t}_{0:N-1}^{pk}\), and likewise a sequence of angular velocity vectors \(\bm{z}_{0:N-1}^{\omega} = \left\{\bm{z}_{i}^{\omega}\right\}_{i=0}^{N-1}\).

\begin{table}[!t]
\vspace{2mm}
	\centering
	\caption{Algorithm: State Decomposition in Unscented Transform}
	\label{table:UT_Decompose}
	\scalebox{1.0}{
		\begin{tabular}{cl}
			\toprule
			\multicolumn{2}{l}{\(\left\{\bm{\mathcal{X}}_{i}, w_{i}^{m}, w_{i}^{c}\right\}_{i=0}^{2d} \leftarrow UTD\left(\bar{\textbf{x}}, \bm{\Sigma}\right)\)}\\
			\midrule
			1: & \hspace{-10pt}
			\(
			\bm{\mathcal{X}}_{0} = \bar{\textbf{x}}
			\)\\
			2: & \hspace{-10pt}
			\(
			\bm{\mathcal{X}}_{i} = \bar{\textbf{x}} + \left(\sqrt{\left(d+\lambda\right)\bm{\Sigma}}\right)_i\ \text{for}\ i=1,\cdots,d
			\)\\
			3: & \hspace{-10pt}
			\(
			\bm{\mathcal{X}}_{i} = \bar{\textbf{x}} - \left(\sqrt{\left(d+\lambda\right)\bm{\Sigma}}\right)_i\ \text{for}\ i=d+1,\cdots,2d
			\)\\
			4: & \hspace{-10pt}
			\(
			w_{0}^{m} = \frac{\lambda}{d+\lambda}
			\)\\
			5: & \hspace{-10pt}
			\(
			w_{0}^{c} = \frac{\lambda}{d+\lambda} + \left(1-\alpha^2+\beta\right)
			\)\\
			6: & \hspace{-10pt}
			\(
			w_{i}^{m} = w_{i}^{c} = \frac{1}{2\left(d+\lambda\right)}\ \text{for}\ i=1,\cdots,2d
			\)\\
			\bottomrule
	\end{tabular}}
\end{table}

\begin{table}[!t]
	\centering
	\caption{Algorithm: State Recovery in Unscented Transform}
	\label{table:UT_Restore}
	\scalebox{1.0}{
		\begin{tabular}{cl}
			\toprule
			\multicolumn{2}{l}{\(\bar{\textbf{x}}, \bm{\Sigma} \leftarrow UTR\left(\left\{\bm{\mathcal{X}}_{i}, w_{i}^{m}, w_{i}^{c}\right\}_{i=0}^{2d}\right)\)}\\
			\midrule
			1: & \hspace{-10pt}
			\(
			\bar{\textbf{x}} = \sum_{i=0}^{2d}{w_{i}^{m} \bm{\mathcal{X}}_{i}}
			\)\\
			2: & \hspace{-10pt}
			\(
			\bm{\Sigma} = \sum_{i=0}^{2d}{w_{i}^{c} \left(\bm{\mathcal{X}}_{i}-\bar{\textbf{x}}\right) \left(\bm{\mathcal{X}}_{i}-\bar{\textbf{x}}\right)^T}
			\)\\
			\bottomrule
	\end{tabular}}
\end{table}

Each \(\bm{z}_{i}^{v}\) is a column vector with linear velocity readings along with \(x\), \(y\), and \(z\) and each \(\bm{z}_{i}^{\omega}\) is a column vector with the angular velocity measurements in \(roll\), \(pitch\), and \(yaw\) in the local frame of reference of the vehicle. In cases where odometry data and lidar packets are asynchronous, \(\bm{z}_{i}^{v}\) and \(\bm{z}_{i}^{\omega}\) can be well approximated using those with the closest timestamps to \(t_{i}^{pk}\), respectively, so long as the assumption that the vehicle kinematic state does not change dramatically during the time difference holds. Also, \(\bm{z}_{i}^{v}\) and \(\bm{z}_{i}^{\omega}\) are assumed to contain independently and identically distributed zero-mean Gaussian noises with their covariance matrices denoted as \(\bm{\Sigma}^{v}\) and \(\bm{\Sigma}^{\omega}\), respectively. The timing jitter in \(t_{i}^{pk}\) is modelled as zero-mean Gaussian noise with its standard deviation \(\sigma_{t}\). Please note that one can choose to use other off-the-shelf ego-motion estimation methods depending on the sensor configurations on the target platforms, as long as the method can provide robust and consistent uncertainty estimates.

Given \(\bm{t}_{0:N-1}^{pk}\), \(\bm{z}_{0:N-1}^{v}\), \(\bm{z}_{0:N-1}^{\omega}\), and \(t_{ref}\), the vehicle ego-motion estimation is required to find out the sequence of Gaussian variables \(\left\{\textbf{x}_{i}^{veh} \sim \mathcal{N}\left(\bar{\textbf{x}}_{i}^{veh}, \bm{\Sigma}_{i}^{veh}\right)\right\}_{i=0}^{N-1}\) representing the estimated vehicle egocentric poses at \(\bm{t}^{lpk}\) with respect to that at \(t_{ref}\). Let \(\textbf{x}_{ref}^{veh}\) denote the Gaussian variable representing the vehicle egocentric state at \(t_{ref}\).
\begin{equation}
\textbf{x}_{ref}^{veh} \sim \mathcal{N}\left(\bar{\textbf{x}}_{ref}^{veh}, \bm{\Sigma}_{ref}^{veh}\right),
\end{equation}
where we set the mean vector \(\bar{\textbf{x}}_{ref}^{veh} = \bm{0}\) and the covariance matrix \(\bm{\Sigma}_{ref}^{veh}\) to a diagonal matrix with close to zero elements, since we are performing ego-motion estimation within the local coordinate system of the vehicle at \(t_{ref}\).

If \(t_{ref} > t_{0}^{lpk}\), then backward ego-motion estimation is performed by first initialising intermediate variables:
\begin{align} \label{eq:egomotion_prediction_init}
t_{*} &\leftarrow t_{ref} & \bar{\textbf{x}}_{*}^{veh} &\leftarrow \bar{\textbf{x}}_{ref}^{veh} & \bm{\Sigma}_{*}^{veh} &\leftarrow \bm{\Sigma}_{ref}^{veh}.
\end{align}

Then for \(i = \max\left( \left\{p : t_{p}^{pk} \in \bm{t}^{pk} \wedge t_{p}^{pk} < t_{ref} \right\} \right), \cdots, 0\), an augmented state vector is constructed by concatenating intermediate vehicle egocentric state \(\textbf{x}_{*}^{veh}\) and kinematic measurements at \(t_{i}^{pk}\).
\begin{equation} \label{eq:egomotion_prediction_start}
\textbf{x}_{*}^{a}
\sim
\mathcal{N}\left(
\bar{\textbf{x}}_{*}^{a},
\bm{\Sigma}_{*}^{a}
\right),
\end{equation}
where
\(
\bar{\textbf{x}}_{*}^{a} =
\begin{bmatrix}
\left(\bar{\textbf{x}}_{*}^{veh}\right)^{T} & \left(\bm{z}_{i}^{v}\right)^{T} & \left(\bm{z}_{i}^{\omega}\right)^{T} & t_{i}^{pk} & t_{*}
\end{bmatrix}
\), and
\(
\bm{\Sigma}_{*}^{a} =
\begin{bmatrix}
\bm{\Sigma}_{*}^{veh} & \bm{0} & \bm{0} & 0 & 0 \\
\bm{0} & \bm{\Sigma}_{v} & \bm{0} & 0 & 0 \\
\bm{0} & \bm{0} & \bm{\Sigma}_{\omega} & 0 & 0 \\
0 & 0 & 0 & \sigma_{t}^{2} & 0 \\
0 & 0 & 0 & 0 & \sigma_{t}^{2} \\
\end{bmatrix}
\).

The backward motion estimation goes from a later timestamp \(t_{*}\) to an earlier \(t_{i}^{pk}\), resulting in a negative time difference considered in the kinematic model.

The augmented state mean and covariance matrix are decomposed through UT into a set of sigma points.
\begin{equation}
\left\{\bm{\mathcal{X}}_{j}^{a}, w_{j}^{m}, w_{j}^{c}\right\}_{j=0}^{2d} \leftarrow UTD\left(\bar{\textbf{x}}_{*}^{a}, \bm{\Sigma}_{*}^{a}\right).
\end{equation}

For \(j = 0,\cdots,2d\), motion estimation is conducted backward in time.
\begin{equation}
\bm{\mathcal{Y}}_{j}^{veh} = f_{km}\left(\bm{\mathcal{X}}_{j}^{a}\right),
\end{equation}
where \(f_{km}\left(\cdot\right)\) is the vehicle kinematic model that predicts vehicle pose based on a given pose and kinematic velocities over a given time duration.

The estimated vehicle egocentric state at timestamp \(t_{i}^{pk}\) is recovered by
\begin{equation}
\bar{\textbf{x}}_{i}^{veh}, \bm{\Sigma}_{i}^{veh} \leftarrow UTR\left(\left\{\bm{\mathcal{Y}}_{j}^{veh}, w_{j}^{m}, w_{j}^{c}\right\}_{j=0}^{2d}\right).
\end{equation}

The results also serve as intermediate variables for the next iteration:
\begin{align} \label{eq:egomotion_prediction_end}
t_{*} &\leftarrow t_{i}^{pk} & \bar{\textbf{x}}_{*}^{veh} &\leftarrow \bar{\textbf{x}}_{i}^{veh} & \bm{\Sigma}_{*}^{veh} &\leftarrow \bm{\Sigma}_{i}^{veh}.
\end{align}

If \(t_{ref} \leq t_{N-1}^{pk}\), then forward vehicle ego-motion estimation is carried out by initialising intermediate variables as in \eqref{eq:egomotion_prediction_init}, and for \(i = \min\left(\left\{p : t_{p}^{pk} \in \bm{t}^{pk} \wedge t_{p}^{pk} \geq t_{ref}\right\}\right),\cdots,N-1\), using the same set of equations \eqref{eq:egomotion_prediction_start}-\eqref{eq:egomotion_prediction_end}, except that in this case
\(
\bar{\textbf{x}}_{*}^{a} =
\begin{bmatrix}
\left(\bar{\textbf{x}}_{*}^{veh}\right)^{T} & \left(\bm{z}_{i}^{v}\right)^{T} & \left(\bm{z}_{i}^{\omega}\right)^{T} & t_{*} & t_{i}^{pk}
\end{bmatrix}
\), and in every iteration the motion estimation starts from an earlier timestamp \(t_{i}^{pk}\) to a later \(t_{*}\).

\subsubsection{3D Lidar Points Motion Correction}

With a sequence of estimated vehicle egocentric poses \(\left\{\textbf{x}_{i}^{veh} \sim \mathcal{N}\left(\bar{\textbf{x}}_{i}^{veh}, \bm{\Sigma}_{i}^{veh}\right)\right\}_{i=0}^{N-1}\) at \(\bm{t}^{pk}\) obtained from the vehicle ego-motion estimation stage, motion correction is applied for each corresponding packet of 3D lidar measurement points.

For \(i = 0,1,\cdots,N-1\), the estimated state mean and covariance matrix are decomposed into a set of sigma points:
\begin{equation}
\left\{\bm{\mathcal{X}}_{i,k}^{veh}, w_{i,k}^{m}, w_{i,k}^{c}\right\}_{k=0}^{2d} \leftarrow UTD\left(\bar{\textbf{x}}_{i}^{veh}, \bm{\Sigma}_{i}^{veh}\right).
\end{equation}

A set of \(4\times 4\) homogeneous transformation matrices \(\left\{\mathcal{T}_{i,k}^{veh}\right\}_{k=0}^{2d}\) are constructed based on the rotation and translation in each \(\bm{\mathcal{X}}_{i,k}^{veh}\).

For \(j = 0,\cdots,M-1\), and for \(k = 0,\cdots,2d\), a motion corrected sigma point is calculated as
\begin{equation}
\bm{\mathcal{Z}}_{i,j,k}^{cld} = (T_{veh}^{ld})^{-1} \cdot \mathcal{T}_{i,k}^{veh} \cdot T_{veh}^{ld} \cdot \bm{z}_{i,j}^{ld},
\end{equation}
where the lidar point is translated to the vehicle's base frame by the rigid transform \(T_{veh}^{ld}\), followed by transformation that encapsulates delta ego-motion in vehicle base frame. Lastly, the point is translated back to the lidar coordinate system.

At this stage, a motion corrected lidar point \(\bm{z}_{i,j}^{cld}\) within lidar packet \(pk_{i}\) can be recovered to a Gaussian variable through
\begin{equation}
\bar{\bm{z}}_{i,j}^{cld}, \bm{\Sigma}_{i,j}^{cld} \leftarrow UTR\left(\left\{\bm{\mathcal{Z}}_{i,j,k}^{cld}, w_{i,k}^{m}, w_{i,k}^{c}\right\}_{k=0}^{2d}\right).
\end{equation}

In the end, the process produces a set of motion corrected sigma points for lidar points denoted by 
\begin{equation}
\bm{\Omega} = \left\{\left\{\left\{\bm{\mathcal{Z}}_{i,j,k}^{cld}\right\}_{j=0}^{M-1}, w_{i,k}^{m}, w_{i,k}^{c}\right\}_{k=0}^{2d}\right\}_{i=0}^{N-1}.
\end{equation}

The lidar ego-motion correction with uncertainty completes at this stage. Further transformation can be applied to \(\bm{\Omega}\) for lidar-to-camera projection with ego-motion uncertainty. Please refer to next section for details.


\subsubsection{Lidar-to-Camera Projection}

This stage is only for systems that require camera-lidar sensor fusion. It takes a motion corrected 3D lidar point cloud as input and project the lidar points to a given camera coordinate system. Essentially, the timestamp of the image for projection has been used as the reference time \(t_{ref}\) in the motion correction process, in pursuance of bringing about an accurate camera-lidar projection. Before projection happens, a 3D lidar point needs to be translated from lidar frame to camera frame given the extrinsic calibration between both camera and lidar sensors represented as the transformation matrix \(T_{cam}^{ld}\):
\begin{equation}
\bm{z}^{cam} = T_{cam}^{ld} \bm{z}^{ld},
\end{equation}
where \(\bm{z}^{cam} = \begin{bmatrix} x^{cam} & y^{cam} & z^{cam} & 1 \end{bmatrix}^{T}\) is the 3D lidar point translated to camera frame.

The generic lidar-to-camera projection function is defined as
\begin{equation}
\begin{bmatrix} u \\ v \end{bmatrix} =
f_{proj}
\left(\bm{z}^{cam}\right),
\end{equation}
which is to find the pixel coordinates \(u\) and \(v\) in the image frame corresponding to a 3D lidar point \(\bm{z}^{cam}\) in the camera frame by using the camera model and its intrinsic parameters.

The function first makes use of the generic pinhole camera-image projection equations, which states
\begin{align}
a &=  \frac{x^{cam}}{z^{cam}} & b &=  \frac{y^{cam}}{z^{cam}} \label{eq_pc1}
\end{align}
\begin{align}
r &= \sqrt{a^{2}+b^{2}} & \theta &= \textup{atan}(r) \label{eq_pc2}.
\end{align}

Since our cameras have fisheye lenses, we need to apply the distortion established by the camera model to find the corresponding pixel in the image \cite{opencv}. The distortion of the lens is calculated as follows:
\begin{equation} \label{eq_pc3}
   \theta_d = \theta(1+k_1\theta^2+k_2\theta^4+k_3\theta^6+k_4\theta^8),
\end{equation}

where $k_1$, $k_2$, $k_3$ and $k_4$ are the lens' distortion coefficients. Then we compute the distorted point coordinates as
\begin{align}
x' &=  (\theta_d/r)a & y' &=  (\theta_d/r)b.
\end{align}

The definite pixel coordinates vector \(\begin{bmatrix} u & v \end{bmatrix}^{T}\) in the image frame of a 3D lidar point can be calculated as
\begin{align}
u &= f_x \cdot (x' + e y')+c_x & v &= f_y \cdot y'+c_y,
\label{eq_pc4}
\end{align}

where \(e\) is the camera's skew coefficient, $[c_x,c_y]$ the principal point offset and $[f_x, f_y]$ are the focal lengths expressed in pixel units.

In order to produce projected points in the image frame with uncertainty information, and also to avoid unnecessary UT, this stage works directly on \(\bm{\Omega}\), which is the set of sigma points for each 3D lidar point as the output of the 3D lidar points motion correction stage.

We can combine translation of each sigma point \(\bm{\mathcal{Z}}_{i,j,k}^{cld} \in \bm{\Omega}\) from lidar frame to camera frame using \(T_{cam}^{ld}\) and the projection to the image frame by
\begin{equation} \label{eq_pc0}
\left\{\bm{\mathcal{K}}_{i,j,k}^{cam} : \left(\exists \bm{\mathcal{Z}}_{i,j,k}^{cld} \in \bm{\Omega}\right)
\left[
\bm{\mathcal{K}}_{i,j,k}^{cam} = f_{proj}\left(T_{cam}^{ld} \bm{\mathcal{Z}}_{i,j,k}^{cld}\right)
\right]
\right\}.
\end{equation}

For \(i = 0,\cdots,N-1\) and for \(j = 0,\cdots,M-1\), the image pixel projected from the lidar point \(\bm{z}_{i,j}^{cld}\) within lidar packet \(pk_{i}\) can be recovered with its mean values and covariance matrix by
\begin{equation}
\begin{bmatrix} \bar{u}_{i,j} \\ \bar{v}_{i,j} \end{bmatrix}, \bm{\Sigma}_{i,j}^{uv} \leftarrow UTR\left(\left\{\bm{\mathcal{K}}_{i,j,k}^{cam}, w_{i,k}^{m}, w_{i,k}^{c}\right\}_{k=0}^{2d}\right).
\end{equation}



\section{Results}
\label{sec:results}

\subsection{Experiment Results}

\begin{figure}[!t]
\vspace{2mm}
\centerline{
\includegraphics[width=0.9\columnwidth]{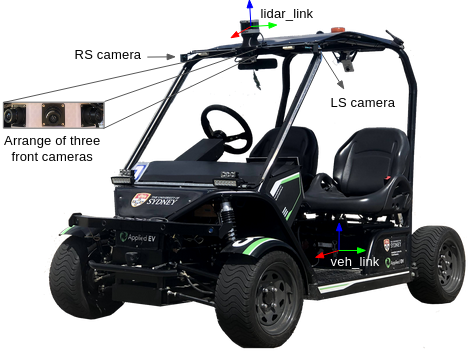}
}
\caption{Experimental platform equipped with five cameras (two side cameras and an arrange of three cameras front facing), one Velodyne VLP-16 lidar, wheel encoders and an IMU that contains gyroscopes, accelerometers and magnetometers. }
\label{fig:platform}
\end{figure}

We implemented the proposed approach in C++ under ROS Melodic release and tested it in the USyd Dataset \cite{usyd_dataset, USYD_Segmentation_2019}, which was obtained with the electric vehicle platform shown in Figure \ref{fig:platform}. The vehicle is equipped with a Velodyne VLP-16 lidar and five fixed lens gigabit multimedia serial link (GMSL) cameras, each covers a \(100^{\circ}\) horizontal field of view. The camera images have a resolution of \(1920 \times 1208\) and are captured at 30 FPS by an onboard NVIDIA DRIVE PX2 automotive computer. The extrinsic camera calibration is conducted relative to the lidar sensor frame, and both are registered to the local frame of reference of the vehicle. The platform also contains wheel encoders and an IMU, which produce odometry measurements at 100 Hz. The constant turn rate and velocity (CTRV) 
kinematic model is adopted for the vehicle.

In the experiment, we use the proposed approach to correct the lidar point cloud using the timestamp of the last lidar packet as \(t_{ref}\). We also project the latest lidar point cloud to the most recent image frames from three front facing cameras, in which case the timestamps of the image frames are chosen as \(t_{ref}\). Only the \(x\) component of the linear velocity measurements is used with the standard deviation of its noise set to 0.05 m/s. The measurements of angular velocities in \(roll\), \(pitch\), and \(yaw\) are used with 2 deg/s set as the standard deviation of their noise, which takes into account the IMU's thermo-mechanical white noise, and the noise from mechanical vibration when the vehicle is moving. \(\sigma_{t}\) is set to 0.0006 s. Each corrected point cloud is published as a ROS \textit{sensor\_msgs/PointCloud2} message, where the data fields of every 3D lidar point are augmented with its covariance in 3D, its projected image coordinates, and their associated covariance in 2D.

\begin{figure}[!t]
\vspace{2mm}
	\centering
	\subfigure[]{ 
		\label{fig:LidarCorrectionLarge:a}
		\includegraphics[trim={1.5cm 0 1.5cm 0.9cm},clip, width=0.9\columnwidth]{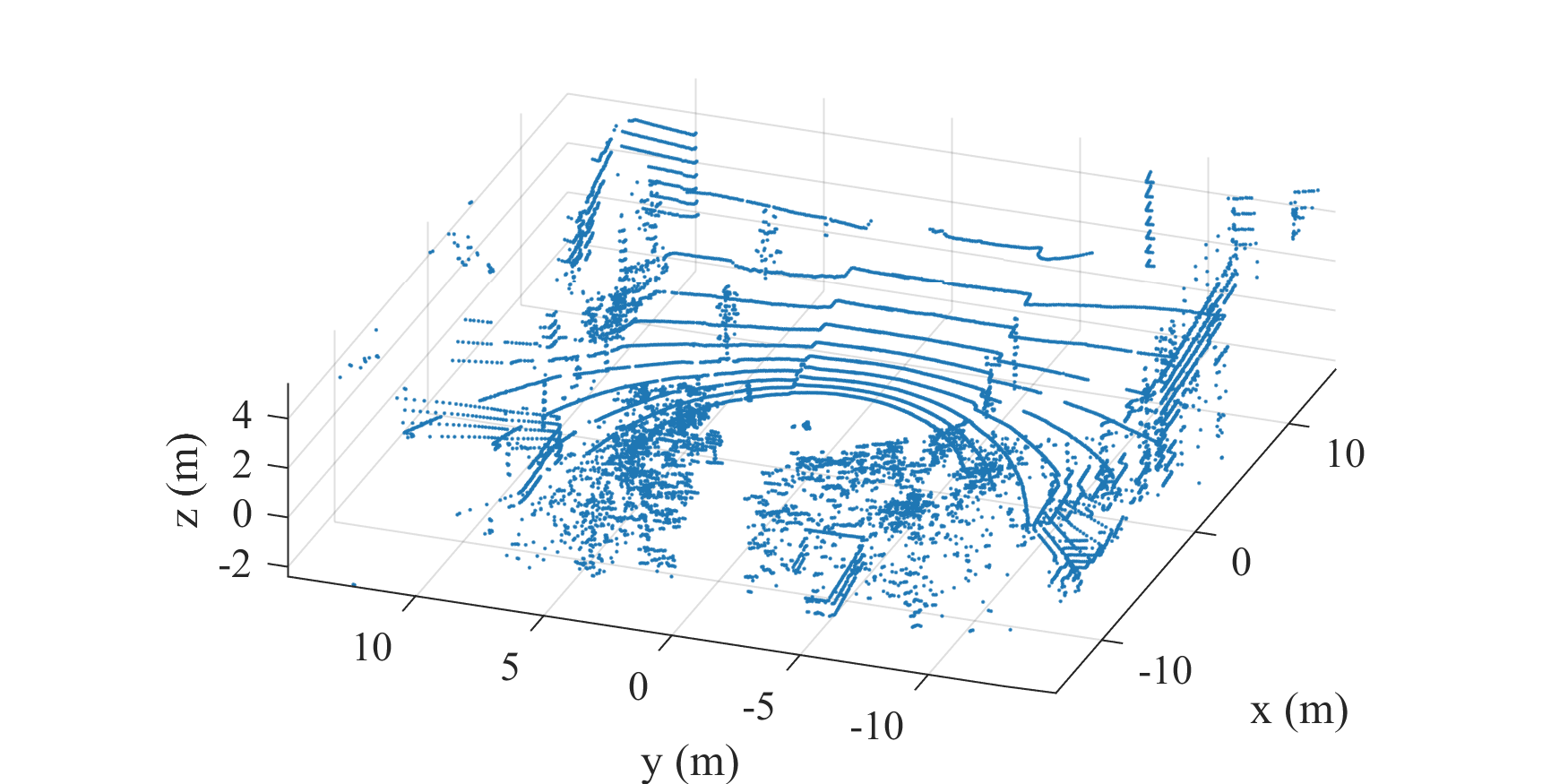}} 
	\hspace{-0.0in}
	\subfigure[]{ 
		\label{fig:LidarCorrectionLarge:b}
		\includegraphics[trim={1.5cm 0 1.5cm 0.9cm},clip, width=0.9\columnwidth]{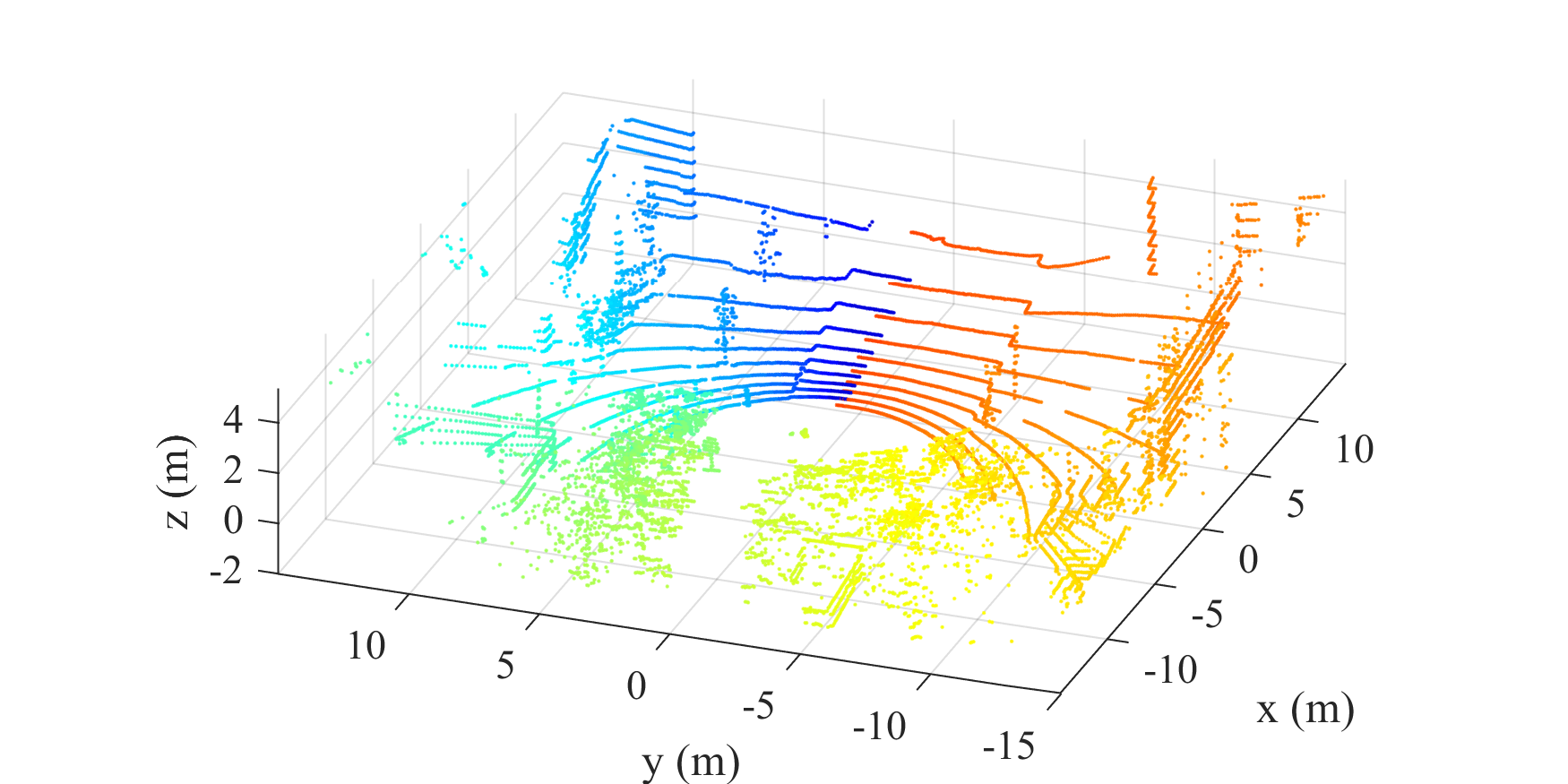}}
	\hspace{-0.0in}
	\subfigure[]{ 
		\label{fig:LidarCorrectionLarge:c}
		\includegraphics[trim={1.5cm 0 1.5cm 0.9cm},clip, width=0.9\columnwidth]{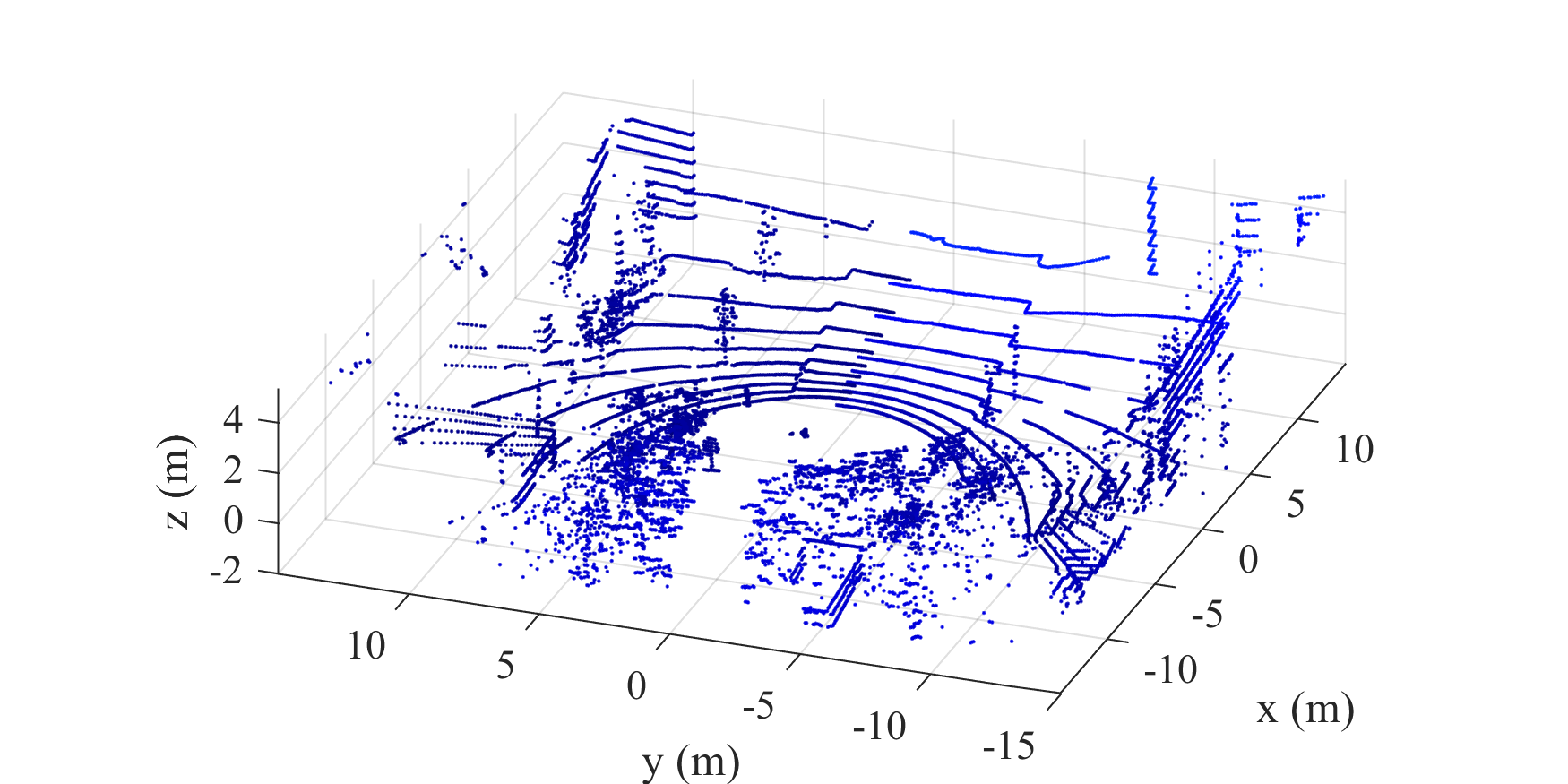}}
	\hspace{-0.0in}
	\subfigure[]{ 
		\label{fig:LidarCorrectionLarge:d}
		\includegraphics[width=2.0in]{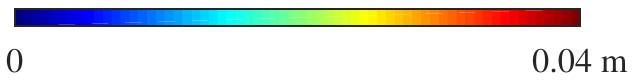}}
	\caption{Lidar point cloud before and after ego-motion correction. (a) shows the point cloud before ego-motion correction. (b) and (c) present the corrected point cloud coloured by the standard deviation in \(x\) and \(y\) directions, respectively. As lidar scans in the clockwise direction, the older points tend to have a higher level of uncertainty due to ego-motion. The uncertainty in \(z\) is found less significant and thus not shown.}
	\label{fig:LidarCorrectionLarge} 
\end{figure}

\begin{figure}[!t]
\vspace{2mm}
	\centering
	\subfigure[]{ 
		\label{fig:LidarCorrectionDetails:a}
		\includegraphics[trim={0 0.15cm  0 0.2cm},clip,width=1.6in]{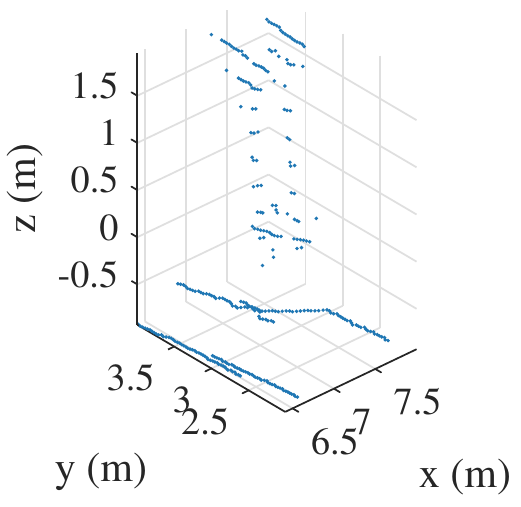}} 
	\hspace{-0.1in}
	\subfigure[]{ 
		\label{fig:LidarCorrectionDetails:b}
		\includegraphics[trim={0 0.15cm  0 0.2cm},clip,width=1.6in]{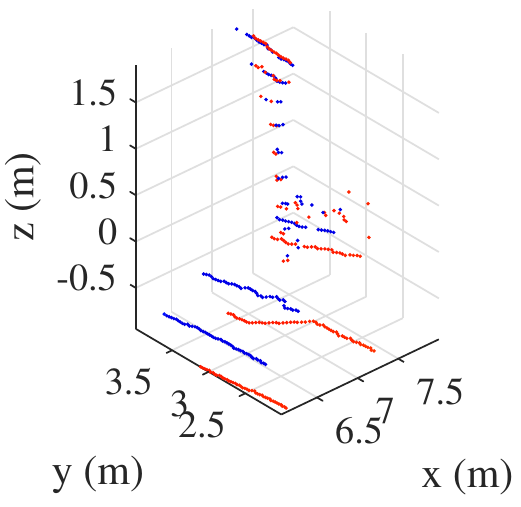}} 
	\subfigure[]{ 
		\label{fig:LidarCorrectionDetails:c}
		\includegraphics[trim={0 0.15cm  0 0.4cm},clip,width=1.3in]{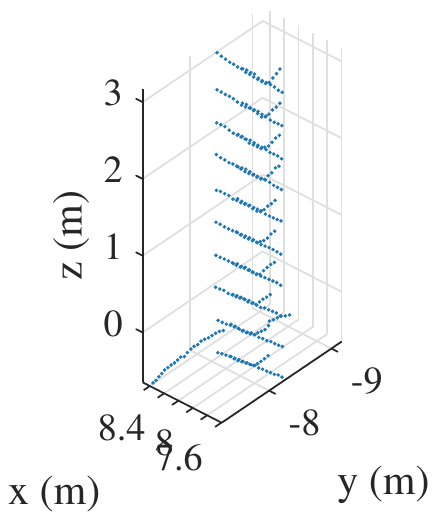}} 
	\hspace{-0.1in}
	\subfigure[]{ 
		\label{fig:LidarCorrectionDetails:d}
		\includegraphics[trim={0 0.15cm  0 0.4cm},clip,width=1.3in]{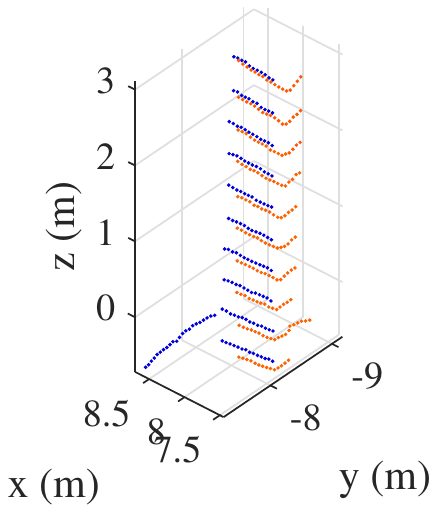}} 
	\subfigure[]{ 
		\label{fig:LidarCorrectionDetails:e}
		\includegraphics[trim={0 0.15cm  0 0.1cm},clip,width=1.6in]{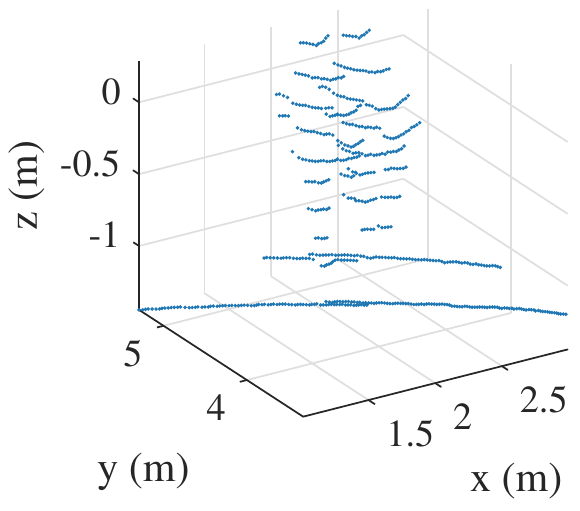}} 
	\hspace{-0.1in}
	\subfigure[]{ 
		\label{fig:LidarCorrectionDetails:f}
		\includegraphics[trim={0 0.15cm  0 0.1cm},clip,width=1.6in]{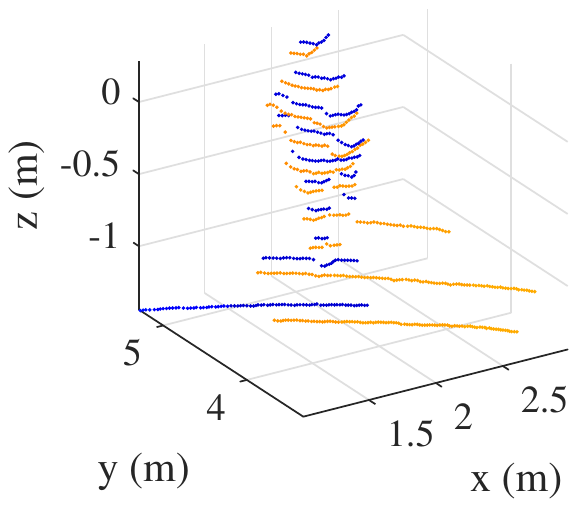}} 
	\hspace{-0.0in}
	\subfigure[]{ 
		\label{fig:LidarCorrectionDetails:g}
		\includegraphics[width=2.0in]{"Figures/colorbar"}}
	\caption{Lidar features before and after ego-motion correction. (a), (c), and (e) illustrate the lidar points of a traffic sign, a pillar, and a pedestrian, respectively, before ego-motion correction. (b), (d), and (f) depict the motion corrected points coloured by the standard deviation in \(x\) direction. The standard deviations in \(y\) and \(z\) directions are available but not shown here.}
	\label{fig:LidarCorrectionDetails} 
\end{figure}

An example of lidar ego-motion correction can be found in Figure \ref{fig:LidarCorrectionLarge}, which clearly demonstrates the corrected lidar point cloud with uncertainty estimates, and is compared with the uncorrected point cloud. The ego-motion distortion can often cause issues in lidar feature extraction. This can, for instance, manifest as a ghost image of the lidar features which is often observed in the overlapping area of the first and last packets of a lidar scan, as shown in the left column of Figure \ref{fig:LidarCorrectionDetails}. As illustrated in the right column of Figure \ref{fig:LidarCorrectionDetails}, the issue can be effectively rectified using the proposed approach, which also provides uncertainty estimate for each lidar point.

\begin{figure}[!t]
\vspace{2mm}
	\centering
	\subfigure[]{ 
		\label{fig:LidarProjectionLarge:a}
		\includegraphics[trim={0.2cm 0.15cm 0.2cm 0.15cm},clip, height=2.37in]{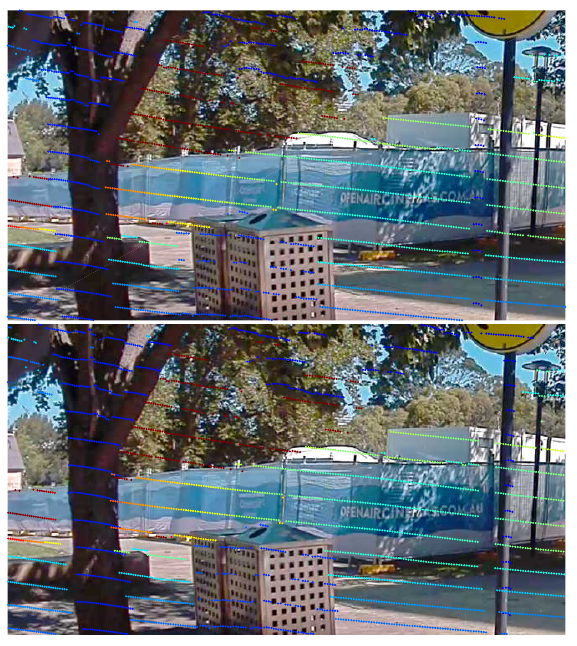}}
	\hspace{-0.13in}
	\subfigure[]{ 
		\label{fig:LidarProjectionLarge:b}
		\includegraphics[trim={0.2cm 0.15cm 0.2cm 0.15cm},clip, height=2.37in]{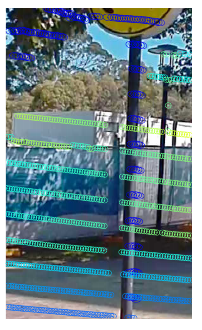}}
	\caption{An example of lidar-to-camera projection using the proposed approach. The projection of the raw point cloud to an image from the front-left camera is shown in top figure of (a), where the misalignment of lidar points and visual features is apparent. A finer overlapping between the points and the image can be observed in the projection of motion corrected lidar points in bottom figure of (a). In addition, (b) shows the uncertainty estimates of each projected point as a result of the proposed approach. Every ellipse in (b) covers a 95\% confidence area. The projected lidar points are colour-coded by range.}
	\label{fig:LidarProjectionLarge} 
\end{figure}


Besides the results previously presented in Figure \ref{fig:LidarProjectionDetails}, more results of lidar-to-camera projection can be found in Figure \ref{fig:LidarProjectionLarge}. It can be clearly seen that precision of the projection improves significantly through the use of the proposed approach. The uncertainty estimates for each projected lidar point on the image are illustrated in Figure \ref{fig:LidarProjectionLarge:b}. It is important to note that in each lidar-to-camera projection figure presented here, the lidar can partially see behind objects seen by the camera. The lidar viewpoint is slightly different to the camera as the sensors are not co-located, and as a result objects observed by one sensor can block the visibility of the other. This is due to the physical setup of cameras and the lidar in the vehicle being mounted in different positions with the aim of providing wide coverage using an array of cameras. In this case, the cameras and lidar perceive the environment from different vantage points. 
Further processing is required to address this occlusion problem \cite{paper:SchneiderHimmelsbach2010}.

\subsection{Simulation Results}

The proposed approach is also assessed quantitatively using simulation, as the ground truth is not available for the experiments with the real vehicle. The simulation is setup as close as possible to the vehicle platform used in the experiment. In every simulation episode the vehicle is moving with ground truth constant linear velocity \(v_{x}\) in the vehicle's \(x\) direction of travel and angular velocity \(\omega_{yaw}\) in \(yaw\) randomly drawn from uniform distributions \(\mathcal{U}\left(2, 10\right)\) m/s and \(\mathcal{U}\left(-60, 60\right)\) deg/s, respectively. As the vehicle moves, the lidar scans for one revolution in 0.1 s, generating 76 lidar packets at different rotational angles. Each packet contains one pair of elevation angle and range data drawn from uniform distributions \(\mathcal{U}\left(-15, 15\right)\) deg and \(\mathcal{U}\left(1, 100\right)\) m, respectively. Linear and angular velocity measurements are polluted with additive Gaussian noise \(\mathcal{N}\left(0, 0.1^{2}\right)\) m/s and \(\mathcal{N}\left(0, 5^{2}\right)\) deg/s, respectively. The timestamp of every piece of sensory measurement contains jitter modelled as Gaussian noise \(\mathcal{N}\left(0, 0.0003^2\right)\) s. Those parameters are chosen to produce a clear result for visualisation. As soon as the lidar finishes one revolution of scanning, the front camera takes an image, to which the lidar point cloud are then projected. Here, the timestamp of the image is used as the reference time. The same intrinsic and extrinsic calibration parameters in the experiment vehicle platform are used in the simulation.

\begin{figure}[!t]
\vspace{2mm}
	\centering
	\subfigure[]{ 
		\label{fig:SimulationLidar:a}
		\includegraphics[width=3.3in]{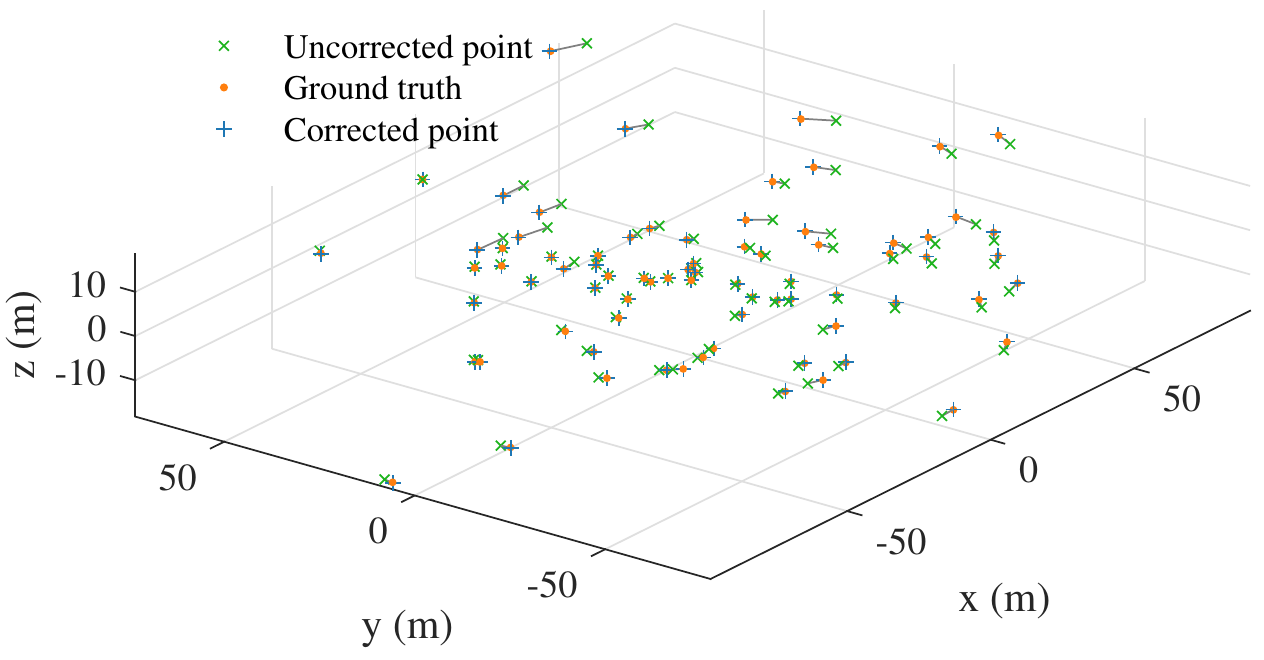}}
	\subfigure[]{ 
		\label{fig:SimulationLidar:b}
		\includegraphics[width=3.1in]{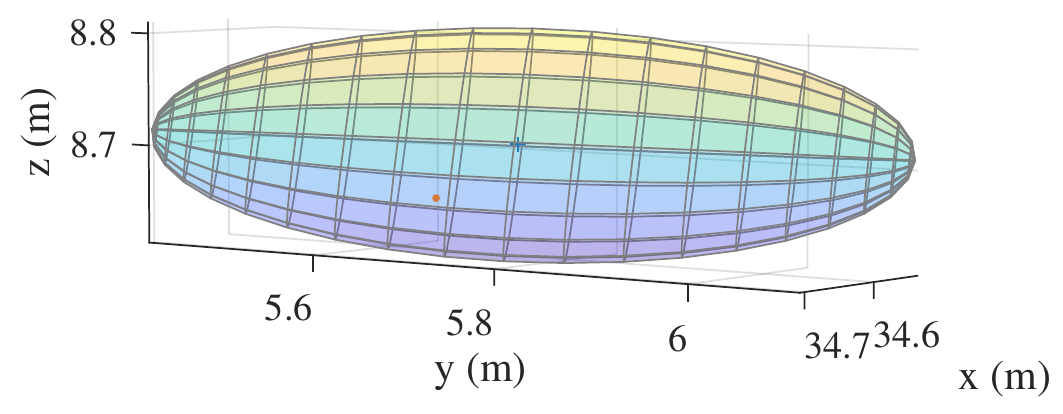}}
	\caption{The comparison of motion corrected lidar point cloud with uncorrected and ground truth point clouds. In (a), as the lidar rotates in clockwise direction, the correction is found more evident to those older points, which have timestamps further from the reference time. The uncertainty of a 3D lidar point after correction is represented as an ellipsoid in (b), which covers 95\% volume of confidence.}
	\label{fig:SimulationLidar} 
\end{figure}

\begin{figure}[!t]
\vspace{2mm}
	\centering
	\subfigure[]{ 
		\label{fig:SimulationImage:a}
		\includegraphics[width=3.1in]{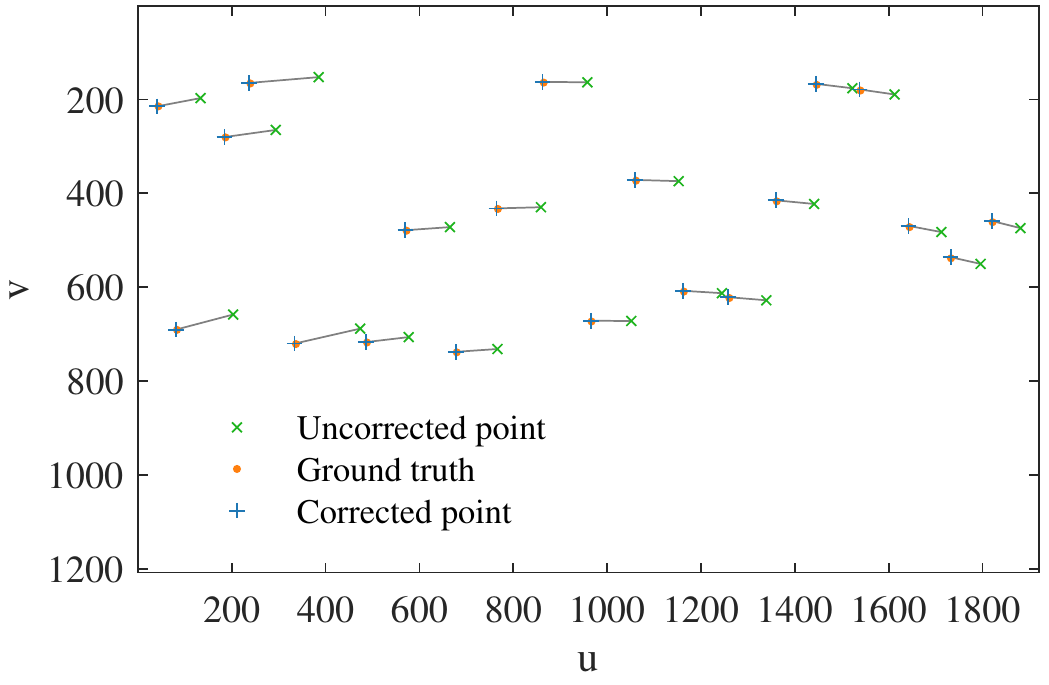}}
	\subfigure[]{ 
		\label{fig:SimulationImage:b}
		\includegraphics[width=3.1in]{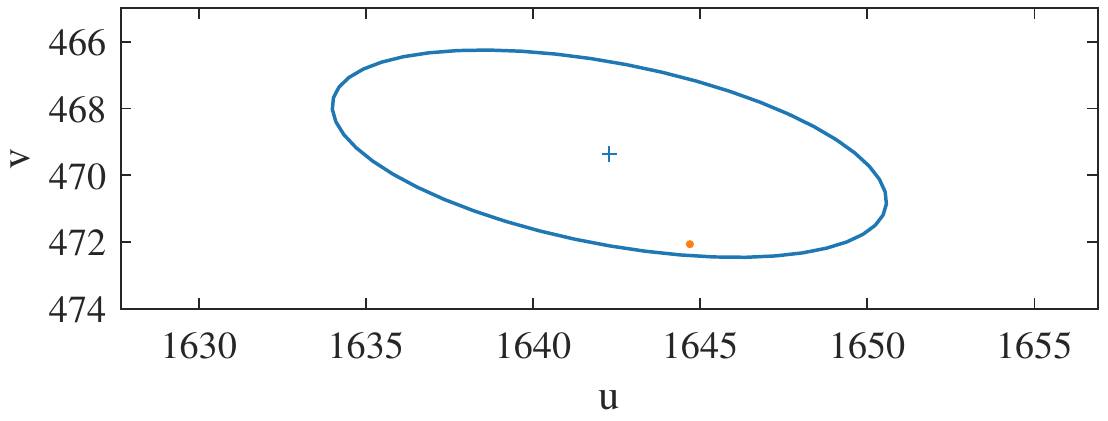}}
	\caption{The projection of motion corrected lidar point cloud to the image compared with that of uncorrected and ground truth point clouds. lidar scans from left to right in the image in (a), where the correction is found more evident to those points at the left side, which have timestamps further from the reference time. The uncertainty of a 2D projected point after correction is represented as an ellipse in (b), which covers a 95\% confidence area.}
	\label{fig:SimulationImage} 
\end{figure}

The simulation results are analysed based on 200 Monte Carlo runs, which in total generate over 15,000 3D lidar points and 4,000 2D projected points. Figure \ref{fig:SimulationLidar} presents the ego-motion corrected lidar point cloud and the comparison with ground truth and uncorrected point clouds from one of the simulation runs, and Figure \ref{fig:SimulationImage} illustrates the same point cloud projected to the image. The ground truth linear and angular velocities in this particular case are 2.81 m/s and -56.2 deg/s (negative means turning right), respectively. Due to the constraint of figure size, we only show the uncertainty of a corrected 3D lidar point and a 2D projected point in Figure \ref{fig:SimulationLidar:b} and Figure \ref{fig:SimulationImage:b}, respectively.

Normalised estimation error squared (NEES) is adopted in the test as the metric of consistency for the proposed lidar ego-motion correction approach. The NEES value for a given 3D lidar sample or 2D projected sample \(\mathcal{N}\left(\bar{\bm{z}}_{i}, \bm{\Sigma}_{i}\right)\) and its ground truth \(\bm{z}_{i}\) is calculated by
\begin{equation}
\epsilon\left(i\right) = \left(\bar{\bm{z}}_{i} - \bm{z}_{i}\right)^{T} \bm{\Sigma}_{i}\left(\bar{\bm{z}}_{i} - \bm{z}_{i}\right).
\end{equation}

\begin{figure}[!t]
	\centering
	\subfigure[]{ 
		\label{fig:NEES:a}
		\includegraphics[trim={0.9cm 0 0.9cm 0.3cm},clip,width=3.2in]{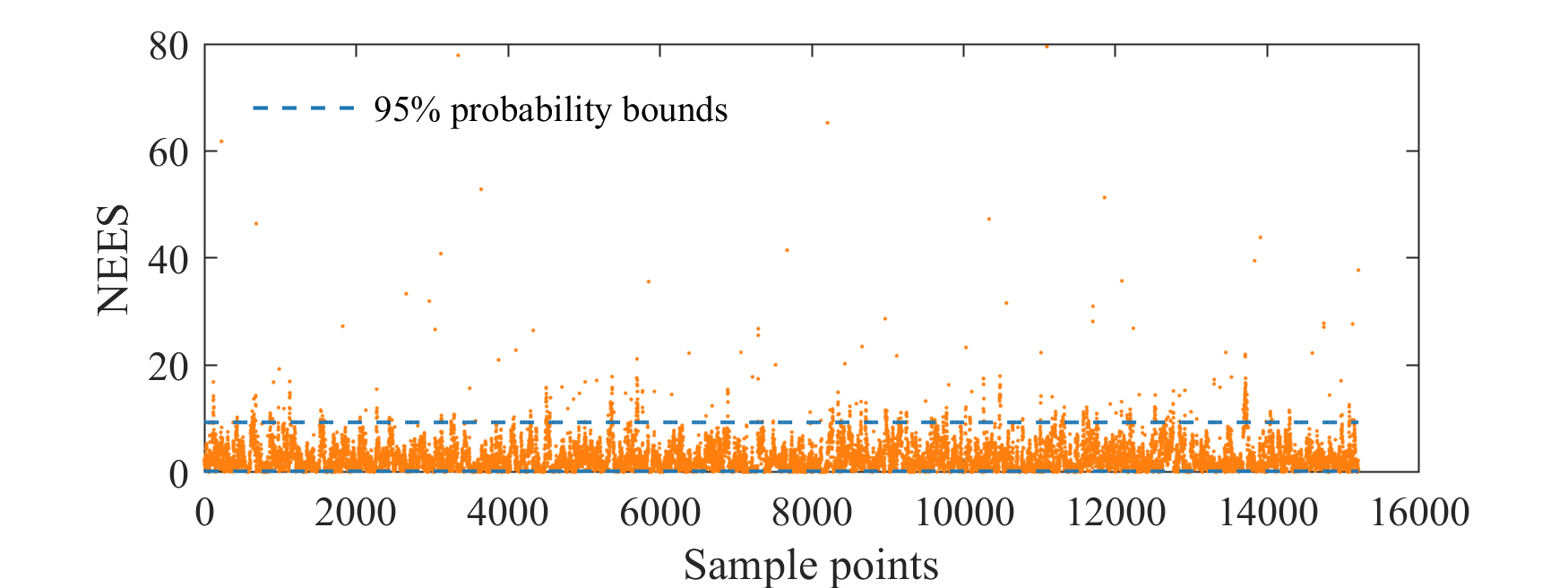}}
	\subfigure[]{ 
		\label{fig:NEES:b}
		\includegraphics[trim={0.9cm 0 0.9cm 0.3 cm},clip,width=3.2in]{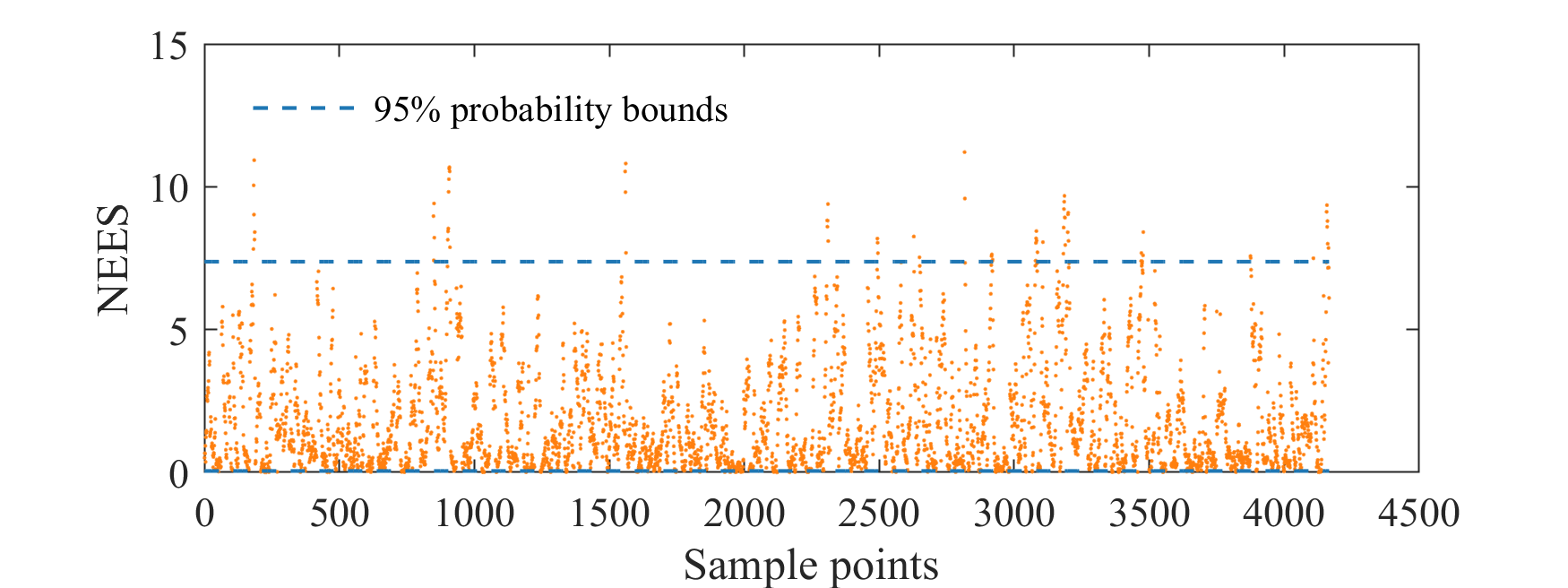}}
	\caption{NEES consistency test for the ego-motion correction results. The in-bound rate for the motion corrected 3D lidar points is about 90.92\%, while the in-bound rate of the projection results in the image frame is found to be 94.29\%. Both results indicate consistent estimation of uncertainties.}
	\label{fig:NEES} 
\end{figure}

Then the \(\epsilon\left(i\right)\) will have a \(\chi^{2}\) (chi-square) distribution with \(dim\left(\bm{z}_{i}\right)\) degrees of freedom, under the hypothesis that the tested estimator is consistent and approximately linear and Gaussian 
\cite{paper:BahrWalter}. The state estimation errors are considered consistent with the calculated covariances if 
\(
\epsilon\left(i\right) \in \left[\chi^{2}_{dim\left(\bm{z}_{i}\right)}\left(0.025\right),\chi^{2}_{dim\left(\bm{z}_{i}\right)}\left(0.975\right)\right]
\), 
where \(dim\left(\bm{z}_{i}\right) = 3\) for a 3D point, and \(dim\left(\bm{z}_{i}\right) = 2\) for a 2D point. This interval associates bounds for the two-sided \(95\%\) probability. The estimation tends to be optimistic if the \(\epsilon\) for all motion corrected lidar points rises significantly higher than the upper bound, while if it stays below the lower bound for a majority of time, the estimator is considered conservative \cite{paper:BaileyNieto}. The consistency check results are presented in Figure \ref{fig:NEES}.

\section{Conclusions And Future Work}
\label{sec:conlusions}

In this paper, a novel probabilistic approach is proposed for lidar ego-motion correction and lidar-to-camera projection with robust uncertainty estimation. The approach accounts for the main error sources which include noise in ego-motion estimation and time jitter in sensor measurements due to practical and theoretical limitations.



The proposed approach considers a sequence of lidar packets, calculates the vehicle ego-motion estimation results for the given packet timestamps, applies the motion correction to the lidar packets against an arbitrarily chosen reference timestamp, and projects the motion corrected lidar points to camera coordinate system. The chain of the above three cascaded stages is formulated into an unscented transform pipeline. Essentially, the corrected and projected points are produced with ego-motion uncertainty information preserved for subsequent processing.

The experimental results demonstrate the accuracy of the ego-motion correction for lidar points, and the projection to the image frame. This was tested on an electric vehicle platform driven in a university campus environment. The simulation results further validate the consistency of the uncertainty estimation in motion correction and lidar-to-camera projection.

Essentially, the capability of producing robust and consistent uncertainty estimates incorporating the lidar ego-motion correction process makes the proposed approach one of the first of its kind to have the potential to be integrated into perception applications that require uncertainty information. The proposed approach associates 3D lidar point and the 2D image coordinates in a probabilistic manner. This is particularly useful in applications that involve probabilistic camera-lidar sensor fusion, where information can be transferred from lidar point to image domain and vice versa with the relevant uncertainty considered.

The future work includes the probabilistic fusion of ego-motion corrected lidar points with semantically labelled images, which combines the heuristic uncertainty associated with a labelled image and the uncertainty from the ego-motion correction of LiDAR point clouds. In this case, the value of the semantic label retrieved from the corresponding pixel in an image frame can be included probabilistically into a point cloud as an additional information field for each 3D point. This helps pave the way to a higher level understanding of the scene, which can be used to enable context based algorithms for collision avoidance and navigation.

\bibliography{main}

\begin{thebibliography}{10}
\providecommand{\url}[1]{#1}
\csname url@samestyle\endcsname
\providecommand{\newblock}{\relax}
\providecommand{\bibinfo}[2]{#2}
\providecommand{\BIBentrySTDinterwordspacing}{\spaceskip=0pt\relax}
\providecommand{\BIBentryALTinterwordstretchfactor}{4}
\providecommand{\BIBentryALTinterwordspacing}{\spaceskip=\fontdimen2\font plus
\BIBentryALTinterwordstretchfactor\fontdimen3\font minus
  \fontdimen4\font\relax}
\providecommand{\BIBforeignlanguage}[2]{{%
\expandafter\ifx\csname l@#1\endcsname\relax
\typeout{** WARNING: IEEEtran.bst: No hyphenation pattern has been}%
\typeout{** loaded for the language `#1'. Using the pattern for}%
\typeout{** the default language instead.}%
\else
\language=\csname l@#1\endcsname
\fi
#2}}
\providecommand{\BIBdecl}{\relax}
\BIBdecl

\bibitem{paper:ChienKlette2016}
H.-J. {Chien}, R.~{Klette}, N.~{Schneider}, and U.~{Franke}, ``Visual odometry
  driven online calibration for monocular {LiDAR}-camera systems,'' in
  \emph{Proc. the 23rd International Conference on Pattern Recognition (ICPR)},
  Dec 2016, pp. 2848--2853.

\bibitem{paper:RiekenMaurer2016}
J.~{Rieken} and M.~{Maurer}, ``Sensor scan timing compensation in environment
  models for automated road vehicles,'' in \emph{Proc. the 19th {IEEE}
  International Conference on Intelligent Transportation Systems (ITSC)}, Nov
  2016, pp. 635--642.

\bibitem{paper:SchneiderHimmelsbach2010}
S.~{Schneider}, M.~{Himmelsbach}, T.~{Luettel}, and H.~{Wuensche}, ``Fusing
  vision and {LIDAR} - {S}ynchronization, correction and occlusion reasoning,''
  in \emph{Proc. {IEEE} Intelligent Vehicles Symposium}, Jun. 2010, pp.
  388--393.

\bibitem{paper:HimmelsbachMuller2010}
M.~Himmelsbach, A.~M\"uller, T.~L\"uttle, and H.-J. W\"unsche, ``{LIDAR}-based
  {3D} object detection,'' in \emph{Proc. the 1st International Workshop on
  Cognition for Technical Systems}, Munich, Germany, Oct. 2008.

\bibitem{paper:MerriauxDupuis2017}
P.~{Merriaux}, Y.~{Dupuis}, R.~{Boutteau}, P.~{Vasseur}, and X.~{Savatier},
  ``{LiDAR point clouds correction acquired from a moving car based on CAN-bus
  data},'' \emph{arXiv e-prints}, p. arXiv:1706.05886, Jun. 2017.

\bibitem{paper:VargaCostea2017}
R.~Varga, A.~Costea, H.~Florea, I.~Giosan, and S.~Nedevschi, ``Super-sensor for
  360-degree environment perception: Point cloud segmentation using image
  features,'' in \emph{Proc. the 20th {IEEE} International Conference on
  Intelligent Transportation Systems (ITSC)}, Oct. 2017.

\bibitem{paper:ByunNa2015}
J.~Byun, K.~i.~Na, B.-S. Seo, and M.~Roh, ``Drivable road detection with 3{D}
  point clouds based on the {MRF} for intelligent vehicle,'' in \emph{Field and
  Service Robotics}.\hskip 1em plus 0.5em minus 0.4em\relax Springer, 2015, pp.
  49--60.

\bibitem{paper:TangYoon2018}
T.~Y. {Tang}, D.~J. {Yoon}, F.~{Pomerleau}, and T.~D. {Barfoot}, ``Learning a
  bias correction for lidar-only motion estimation,'' in \emph{Proc. the 15th
  Conference on Computer and Robot Vision (CRV)}, May 2018, pp. 166--173.

\bibitem{paper:MoosmannStiller2011}
\BIBentryALTinterwordspacing
F.~Moosmann and C.~Stiller, ``\BIBforeignlanguage{english}{Velodyne {SLAM}},''
  in \emph{\BIBforeignlanguage{english}{Proc. the {IEEE} Intelligent Vehicles
  Symposium}}, Baden-Baden, Germany, Jun. 2011, pp. 393--398. [Online].
  Available:
  \url{\url{http://www.mrt.kit.edu/z/publ/download/Moosmann\_IV11.pdf}}
\BIBentrySTDinterwordspacing

\bibitem{paper:ZhangSingh2014}
J.~Zhang and S.~Singh, ``{LOAM}: {L}idar odometry and mapping in real-time,''
  in \emph{Proc. Robotics: Science and Systems Conference}, Jul. 2014.

\bibitem{paper:HongKo2010}
S.~Hong, H.~Ko, and J.~Kim, ``{VICP}: {V}elocity updating iterative closest
  point algorithm,'' in \emph{Proc. {IEEE} International Conference on Robotics
  and Automation ({ICRA})}, Anchorage, Alaska, USA, May 2010, pp. 1893--1898.

\bibitem{paper:LeGentilVidalCalleja2018}
C.~{Le Gentil}, T.~{Vidal-Calleja}, and S.~{Huang}, ``{3D} lidar-{IMU}
  calibration based on upsampled preintegrated measurements for motion
  distortion correction,'' in \emph{Proc. {IEEE} International Conference on
  Robotics and Automation ({ICRA})}, Brisbane, Australia, May 2018, pp.
  2149--2155.

\bibitem{paper:CharikaShan2019}
C.~{De Alvis}, M.~{Shan}, S.~{Worrall}, and E.~{Nebot}, ``Uncertainty
  estimation for projecting lidar points onto camera images for moving
  platforms,'' in \emph{Proc. IEEE International Conference on Robotics and
  Automation (ICRA)}, May 2019, pp. 6637--6643.

\bibitem{ros_camera_calib}
J.~Bowman and P.~Mihelich, ``{ROS} perception: camera\_calibration,''
  \url{http://wiki.ros.org/camera\_calibration}, 2020, accessed: 2020-03-01.

\bibitem{fish_eye_model}
J.~{Kannala} and S.~S. {Brandt}, ``A generic camera model and calibration
  method for conventional, wide-angle, and fish-eye lenses,'' \emph{IEEE
  Transactions on Pattern Analysis and Machine Intelligence}, vol.~28, no.~8,
  pp. 1335--1340, Aug. 2006.

\bibitem{SurabhiITSC}
S.~Verma, J.~S. Berrio, S.~Worrall, and E.~Nebot, ``Automatic extrinsic
  calibration between a camera and a {3D} lidar using {3D} point and plane
  correspondences,'' in \emph{Proc. {IEEE} International Conference on
  Intelligent Transportation Systems ({ITSC})}, Auckland, New Zealand, Oct.
  2019, pp. 3906--3912.

\bibitem{opencv}
Open\_CV, ``Fisheye camera model, camera calibration and {3D} reconstruction,''
  \url{https://docs.opencv.org/3.4/db/d58/group\_\_calib3d\_\_fisheye.html},
  2020, accessed: 2019-03-01.

\bibitem{usyd_dataset}
\BIBentryALTinterwordspacing
W.~Zhou, S.~Berrio, S.~Worrall, C.~{De Alvis}, M.~Shan, J.~Ward, and E.~Nebot,
  ``The usyd campus dataset,'' 2019. [Online]. Available:
  \url{http://dx.doi.org/10.21227/sk74-7419}
\BIBentrySTDinterwordspacing

\bibitem{USYD_Segmentation_2019}
\BIBentryALTinterwordspacing
W.~Zhou, S.~Berrio, S.~Worrall, and E.~Nebot, ``Automated evaluation of
  semantic segmentation robustness for autonomous driving,'' in \emph{IEEE
  Transactions on Intelligent Transportation Systems}, Early Print, April 2019.
  [Online]. Available: \url{https://doi.org/10.1109/TITS.2019.2909066}
\BIBentrySTDinterwordspacing

\bibitem{paper:BahrWalter}
A.~Bahr, M.~R. Walter, and J.~J. Leonard, ``Consistent cooperative
  localisation,'' in \emph{Proc. {IEEE} International Conference on Robotics
  and Automation ({ICRA})}, Kobe, Japan, May 2009.

\bibitem{paper:BaileyNieto}
T.~Bailey, J.~Nieto, J.~Guivant, M.~Stevens, and E.~Nebot, ``Consistency of the
  {EKF-SLAM} algorithm,'' in \emph{Proc. {IEEE/RSJ} International Conference of
  Intelligent Robots and Systems ({IROS})}, Beijing, China, Oct. 2006.

\end{thebibliography}
\bibliographystyle{IEEEtran}

\end{document}